\documentclass[lettersize,journal]{IEEEtran}

\usepackage{bbding}

\usepackage{array}
\usepackage[caption=false,font=normalsize,labelfont=sf,textfont=sf]{subfig}
\usepackage{textcomp}
\usepackage{stfloats}
\usepackage{url}
\usepackage{verbatim}
\usepackage{graphicx}
\usepackage{cite}
\usepackage{epsfig}
\usepackage{amssymb}
\usepackage{multirow}
\usepackage{booktabs}
\usepackage{color}
\usepackage{footnote}
\usepackage{engord}

\usepackage{lipsum}

\newtheorem{definition}{Definition}

\usepackage{algorithm}
\usepackage{algorithmic}
\usepackage{amsmath}
\usepackage{amsfonts}
\usepackage{mathtools}
\usepackage{booktabs}

\definecolor{Grey}{rgb}{0.0,0.1,0.5}

\newcommand{\revise}[1]{{\color{black}#1}}

\newcommand{\minorrevise}[1]{{\color{black}#1}}

\newcommand{\executeiffilenewer}[3]{%
	\ifnum\pdfstrcmp{\pdffilemoddate{#1}}%
	{\pdffilemoddate{#2}}>0%
	{\immediate\write18{#3}}\fi%
}

\graphicspath{{./}{section/figure/}{images/}}

\begin{document}

\title{Transferability-Guided Cross-Domain Cross-Task Transfer Learning}

\author{Yang~Tan, Enming~Zhang, Yang~Li, Shao-Lun Huang, Xiao-Ping Zhang,~\IEEEmembership{Fellow,~IEEE}

\thanks{This work has been accepted by IEEE TNNLS. Copyright may be transferred without notice, after which this version may no longer be accessible. } 
\thanks{Yang Tan, Enming Zhang, Yang Li, Shao-Lun Huang and Xiao-Ping Zhang are with Shenzhen Key Laboratory of Ubiquitous Data Enabling, Shenzhen International Graduate School, Tsinghua University. The corresponding author is Yang Li (e-mail: yangli@sz.tsinghua.edu.cn).}}


\maketitle

\begin{abstract}

We propose two novel transferability metrics F-OTCE (Fast Optimal Transport based Conditional Entropy) and JC-OTCE (Joint Correspondence OTCE) to evaluate how much the source model (task) can benefit the learning of the target task and to learn more generalizable representations for cross-domain cross-task transfer learning. Unlike the original OTCE metric that requires evaluating the empirical transferability on auxiliary tasks, our metrics are auxiliary-free such that they can be computed much more efficiently. Specifically, F-OTCE estimates transferability by first solving an Optimal Transport (OT) problem between source and target distributions, and then uses the optimal coupling to compute the Negative Conditional Entropy between source and target labels. \revise{It can also serve as an objective function to enhance downstream transfer learning tasks including model finetuning and domain generalization}. Meanwhile, JC-OTCE improves the \revise{transferability accuracy} of F-OTCE by including label distances in the OT problem, though it incurs additional computation costs. Extensive experiments demonstrate that F-OTCE and JC-OTCE outperform state-of-the-art auxiliary-free metrics by $21.1\%$ and $25.8\%$, respectively in correlation coefficient with the ground-truth transfer accuracy. By eliminating the training cost of auxiliary tasks, the two metrics reduce the total computation time of the previous method from 43 minutes to 9.32s and 10.78s, respectively, for a pair of tasks. \revise{When applied in the model finetuning and domain generalization tasks, F-OTCE shows significant improvements in the transfer accuracy in few-shot classification experiments, with up to $4.41\%$ and $2.34\%$ accuracy gains, respectively.}

\end{abstract}


\section{Introduction}

\IEEEPARstart{T}{ransfer} learning is an effective learning paradigm to enhance the performance on target tasks via leveraging prior knowledge from the related source tasks (or source models), \revise{especially when there are only few labeled data for supervision~\cite{pratt1993discriminability,sun2019meta, shao2014transfer, zhang2022transfer}}. However, the success of transfer learning is not always guaranteed. If the source and target tasks are unrelated, or if the transferred representation does not carry sufficient information about the target task, transfer learning will not obtain a notable gain on the target task performance, and may even experience negative transfer, i.e., the performance becomes worse than that of training from scratch on the target task~\cite{zhang2020overcoming}.  Therefore, understanding when and what to transfer between tasks is crucial to the success of transfer learning.\par 

\begin{figure}[t]
	\centering
	\includegraphics[width=0.9\linewidth]{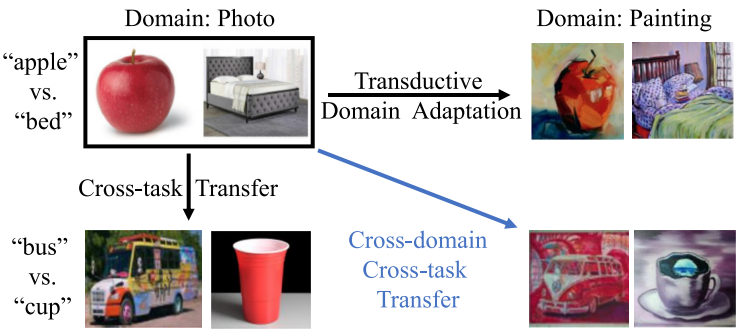}
	\caption{Illustration of three different transfer learning settings, i.e., transductive domain adaptation~\cite{pan2009survey}, cross-task transfer~\cite{bao2019information} and the cross-domain cross-task transfer we investigating.}
	\label{fig:illustration of cross domain cross task transfer}
\end{figure}

The ``when to transfer'' problem was traditionally studied theoretically through the derivation of generalization bounds of transfer learning across tasks~\cite{maurer2009transfer, ben2003exploiting} and \revise{across domains (also known as the domain adaptation problem)~\cite{ben2010theory, ben2006analysis, blitzer2008learning,mansour2009domain, wang2021rethinking, wang2019domain}}. Such studies bound the target task generalization error by a function that depends on certain divergence between the source and target domain, or the complexity of the hypothesis class for the source and target tasks. In practice, however, these bounds are difficult to compute from data and 
they tend to rely on strict assumptions that can not be verified. In recent years, the notion of task transferability was proposed to address the ``when to transfer'' problem in the context of deep transfer learning~\cite{zamir2018taskonomy, yosinski2014transferable, achille2019task2vec, ying2018transfer, NCE,bao2019information,LEEP,OTCE,LogME}. The transferability problem aims to quantitatively evaluate how much the source task or source model could benefit the learning of the target task. It can be used to directly select the most ``transferable" source model from a model zoo for a target task, rather than exhaustively trying  each source model on the target data. In addition, transferability can help prioritize different tasks for joint training~\cite{zamir2018taskonomy} and multi-source feature fusion~\cite{OTCE}.\par

As ~\revise{empirical transferability studies~\cite{zamir2018taskonomy, yosinski2014transferable, achille2019task2vec, ying2018transfer, huang2022balancing}} incur heavy computation burdens in retraining the transfer learning model on the target training data, a new trend of transferability research aims to efficiently estimate the transfer performance a-priori with little or no training of the transfer model. Several efficient transferability metrics are proposed, including NCE~\cite{NCE}, H-score~\cite{bao2019information}, LEEP~\cite{LEEP}, and LogME~\cite{LogME}. Despite being evidently more efficient to compute from practical data than empirical methods, they are also prone to strict data assumptions \cite{NCE, bao2019information} and insufficient performance \cite{LEEP, LogME} while task complexities are similar. \revise{Moreover, the aforementioned metrics are solely used for determining when to transfer between a pair of source and target tasks, but they do not contribute to solving the ``what to transfer" problem, i.e. how to obtain more generalizable feature representations across domains and tasks.}\par

Recently, a novel transferability metric \textbf{OTCE} (Optimal Transport based Conditional Entropy)~\cite{OTCE} is proposed to effectively estimate the transferability under the challenging cross-domain cross-task transfer setting, as shown in Fig. \ref{fig:illustration of cross domain cross task transfer}. \revise{Unlike the transferability metrics mentioned earlier, OTCE adopts a more analytical disentanglement approach. It explicitly assesses the domain difference (measured by Wasserstein distance) and the task difference (determined by conditional entropy) between tasks, and then integrates them via a linear model to quantify transferability. This technically sound design yields substantial accuracy improvement over the aforementioned metrics.} Nevertheless, a major limitation of OTCE is its dependency on auxiliary tasks with known transfer performance to determine the intrinsic parameters of the linear model. \minorrevise{On one hand, the availability of sufficient labeled data for creating auxiliary tasks is not always guaranteed. On the other hand, assessing the transfer performance of such auxiliary tasks necessitates retraining the source model, incurring additional computation costs.} As a result, the reliance on auxiliary tasks makes OTCE relatively inefficient and less applicable in general practical scenarios.\par

 \revise{In this paper, we aim to broaden the applicability of the OTCE framework and investigate the potential uses of transferability in downstream transfer learning tasks.} We propose two auxiliary-free transferability metrics, namely \textbf{F-OTCE} (Fast OTCE) and \textbf{JC-OTCE} (Joint Correspondence OTCE), which eliminate the need for auxiliary tasks and substantially enhance the efficiency without compromising accuracy. For classification problems, the F-OTCE metric solves the Optimal Transport (OT) problem~\cite{kantorovich1942translocation,peyre2019computational} to estimate a probabilistic coupling between the unpaired samples from the source and target datasets. Then the optimal coupling enables us to derive the negative conditional entropy between the source and target task labels for representing transferability, which measures the label uncertainty of a target sample given the labels of corresponding source samples. While the F-OTCE metric does not explicitly evaluate the domain difference, the estimated probabilistic coupling between the source and target data implicitly captures the domain difference to some extent in this unified framework.\par

\revise{Then we propose the JC-OTCE metric to further improve the accuracy of the F-OTCE metric in diverse transfer configurations. Our motivation is that F-OTCE only considers the joint probability distribution of input samples when determining data correspondences between the source and target domains. But this approach is limited because the definition (label annotations) of the source task can also affect model generalization. To address this limitation, we incorporate label distance into the ground cost of the OT problem, allowing for the computation of correspondences in both sample and label spaces. By including additional label information, JC-OTCE produces improved data correspondences that partially compensate for the lack of explicit domain difference consideration. JC-OTCE achieves comparable transferability accuracy to the original OTCE metric but requires additional computation compared to F-OTCE, which remains preferable for efficiency purposes.}

\revise{Moreover, we investigate the application of our transferability metric in two downstream transfer learning tasks including \textit{model finetuning} and \textit{domain generalization}, offering a solution to the ``what to transfer'' problem. Specifically, to enhance the model finetuning performance, we propose an OTCE-based finetune algorithm that optimizes the pretrained source model to learn more transferable feature representation via maximizing the F-OTCE score between the source and target tasks. The optimized model is then finetuned on target training data using the classification loss function.} 

\revise{We also demonstrate that incorporating the F-OTCE metric into a novel domain generalization method URL~\cite{li2021universal} can further improve its generalizability on unseen domains. Our motivation is to view distilling knowledge from domain-specific models to the universal model as maximizing the transferability between them. Therefore, we replace the knowledge distillation function in URL with our F-OTCE score, resulting in significant accuracy improvements in few-shot classification tasks on unseen domains. }

\revise{This work is an extension of our previous conference paper~\cite{OTCE}, and the additional contributions are summarized as follows:\par

\begin{enumerate}[]
	\item \textit{Expanding the applicability of OTCE framework}. Our proposed F-OTCE and JC-OTCE metrics eliminate the need for auxiliary tasks and achieve comparable transferability accuracy to OTCE. They also outperform previous auxiliary-free transferability metrics in terms of accuracy while maintaining comparable efficiency.

	\item \textit{Investigating the potential uses of transferability.} We illustrate the effectiveness of using F-OTCE as an optimization objective in improving the performance of downstream tasks, such as model finetuning and domain generalization. We consider F-OTCE to be a general tool that can be easily integrated into various algorithms for transfer learning and other related applications.

\end{enumerate}

}

In our experiments using several multi-domain classification datasets, we show that our proposed two metrics significantly outperform existing auxiliary-free metrics with $25.8\%$ correlation gain on average, while cutting more than $99\%$ of the computation time in the original OTCE. We also show that, when served as a loss function, F-OTCE leads to notable classification accuracy gains on the model finetuning and domain generalization tasks, with up to $4.41\%$ and $2.34\%$. The rest of this paper is organized as follows. Section \ref{sec: preliminary} introduces the formulation of transferability. Section \ref{sec: prelimianry of OTCE} provides a preliminary analysis on OTCE. Section \ref{sec: two auxiliary metrics} presents our two auxiliary-free transferability metrics. Section \ref{sec: otce-based finetune} illustrates our proposed transferability-guided model finetuning and domain generalization algorithms. Section \ref{sec: experiment} provides all the experimental results and analyses. Finally, we draw the conclusion in Section \ref{sec: conclusion}.

\section{Transferability Formulation}
\label{sec: preliminary}

\begin{figure*}[t]
	\centering
	\includegraphics[width=0.9\linewidth]{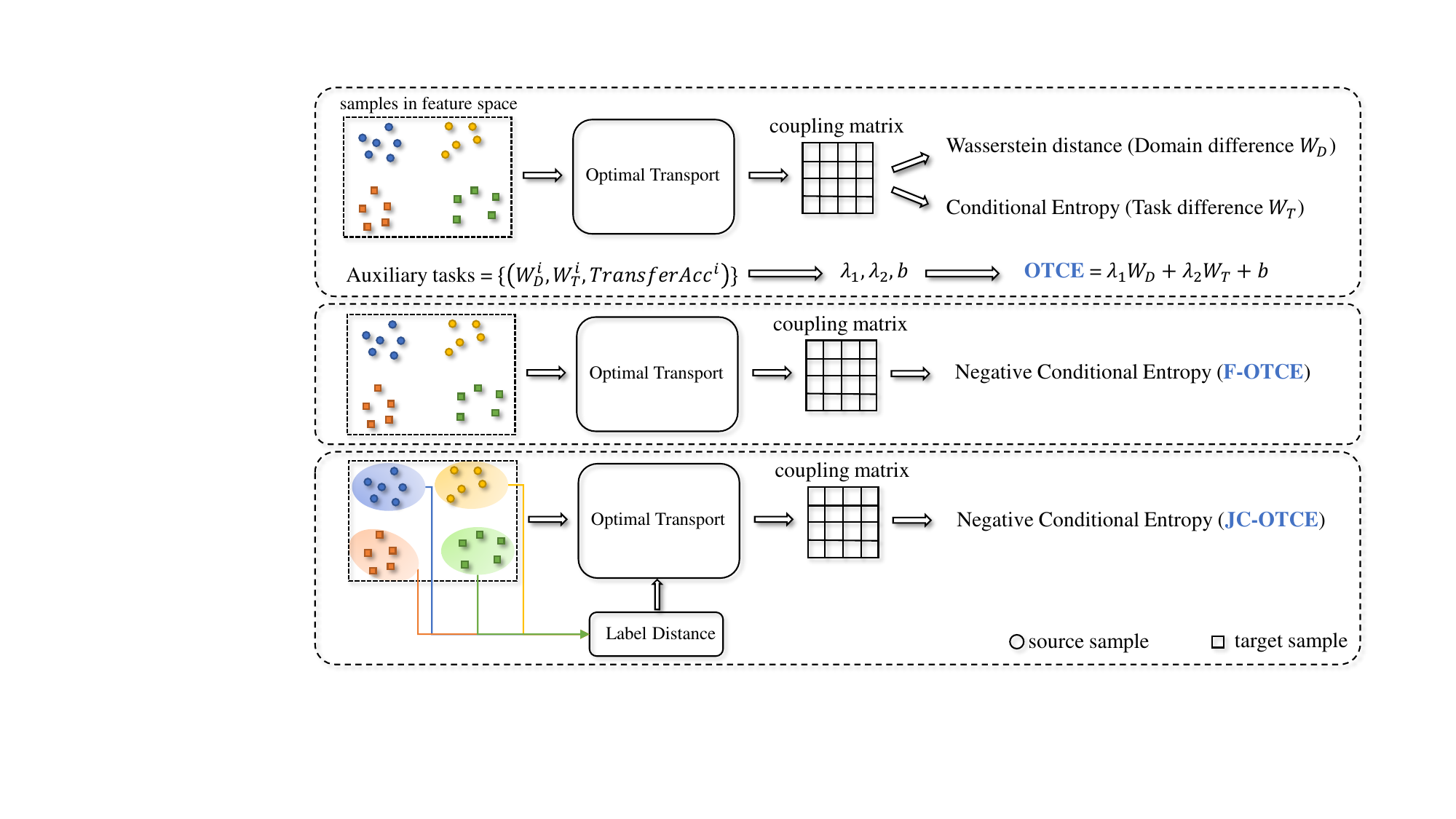}
	\caption{Illustration of the auxiliary-based OTCE metric~\cite{OTCE} (top), and our proposed F-OTCE (middle) and JC-OTCE (bottom) metrics which do not require auxiliary tasks with known transfer accuracy to learn the weighting coefficients. For OTCE (top), $W_D$ and $W_T$ represent the domain difference and task difference between two tasks, respectively. To estimate the coefficients $\lambda_1, \lambda_2, b$ of the linear model,  
	we need to sample at least three auxiliary tasks from the target dataset and calculate $W_D^i$, $W_T^i$ and transfer accuracy $TransferAcc^i$ between the source task and each auxiliary task as training data.
	}
	\label{fig:otce framework comparison}
\end{figure*}

Here we introduce the formal definition of transferability. Suppose we have source data $D_s = \{(x^i_s, y^i_s)\}_{i=1}^m \sim P_s(x,y)$ and target data $D_t = \{(x^i_t, y^i_t)\}_{i=1}^n \sim P_t(x,y)$, where $x$ represents the input instance and $y$ denotes the label. We have $x^i_s, x^i_t$ from the input space $\mathcal{X}$, and $ y^i_s$ from the source label space $\mathcal{Y}_s$, and $ y^i_t$ from the target label space $\mathcal{Y}_t$. Meanwhile, $P(x_s) \neq P(x_t)$ and $\mathcal{Y}_s \neq \mathcal{Y}_t$ indicate different domains and tasks respectively. In addition, we are given a source model $(\theta_s, h_s)$ pretrained on source data $D_s$, in which $\theta_s: \mathcal{X}\rightarrow\mathbb{R}^d$ represents a feature extractor producing $d$-dimensional features and $h_s:\mathbb{R}^d \rightarrow \mathcal{P}(\mathcal{Y}_s)$ is the head classifier predicting the final probability distribution of labels, where $\mathcal{P}(\mathcal{Y}_s)$ is the space of all probability distributions over $\mathcal{Y}_s$. Note that the notation $\theta$ and $h$ can also represent model parameters.

In this paper, we mainly investigate the transferability estimation problem with two representative transfer paradigms for neural networks~\cite{zhang2020overcoming}, i.e., \textit{Retrain head} and \textit{Finetune}. The \textit{Retrain head} method keeps the parameters of the source feature extractor $\theta_s$ frozen and retrains a new head classifier $h_t$. But the \textit{Finetune} method updates the source feature extractor and the head classifier simultaneously to obtain new ($\theta_t, h_t$). Compared to \textit{Retrain head}, \textit{Finetune} trade-offs transfer efficiency for better transfer accuracy and it requires more target data to avoid overfitting~\cite{OTCE}.

To obtain the empirical transferability, we need to retrain the source model via Retrain head or Finetune on target data and then evaluate the expected log-likelihood on its testing set. Formally, the empirical transferability is defined as:
\begin{definition}
	The empirical transferability from the source task $S$ to the target task $T$ is measured by the expected log-likelihood of the retrained $(\theta_s, h_t)$ or $(\theta_t, h_t)$ on the testing set of target task:
	\begin{equation}
		\mathrm{Trf}(S \rightarrow T) = 
		\begin{cases}
			\mathbb{E} \left[ \mathrm{log}\ P(y_t|x_t; \theta_s, h_t) \right] & \text{(Retrain head)}\\
			\mathbb{E} \left[ \mathrm{log}\ P(y_t|x_t; \theta_t, h_t) \right] & \text{(Finetune)}\\
		\end{cases},
		\label{eq:true-transferability}
	\end{equation}
	which indicates how good the transfer performance is on the target task. In practice, we usually take the testing accuracy as an approximation of the log-likelihood~\cite{NCE, OTCE}. 
\end{definition}\par

Although the empirical transferability can be the golden standard of describing how easy it is to transfer the knowledge learned from a source task to a target task, it is computationally expensive to obtain. Efficient transferability metric is a function of the source and target data that approximates the empirical transferability, i.e., the \textit{ground-truth} of the transfer performance on target tasks. It is therefore imperative to find efficient transferability metrics that can accurately estimate empirical transferability.

 \section{\minorrevise{Preliminary Analysis of OTCE}}
 \label{sec: prelimianry of OTCE}
 
\begin{figure}[t]
	\centering
	\includegraphics[width=1.0\linewidth]{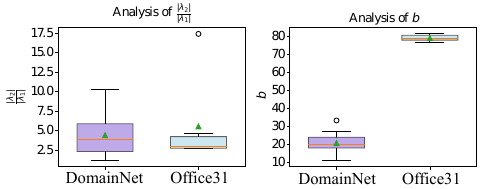}
	\caption{Statistic of the learned weighting coefficients $\lambda_1, \lambda_2$ and the bias term $b$ of OTCE under diverse transfer configurations. }
	\label{fig: coefficients analysis}
\end{figure}

\minorrevise{OTCE (Optimal Transport based Conditional Entropy) is an analytical transferability metric proposed for the cross-domain cross-task transfer learning setting. As illustrated in Fig. \ref{fig:otce framework comparison} (upper part), OTCE quantifies transferability as a linear model of the domain difference $W_D$ (measured by Wasserstein distance) and task difference $W_T$ (determined by conditional entropy), which is denoted as:
	\begin{equation}
	\text{OTCE} = \lambda_1 W_D + \lambda_2 W_T + b.
\end{equation}

However, a major limitation of OTCE is its dependency on auxiliary tasks with known transfer accuracy to determine the intrinsic parameters of the linear model. In practice, we are not always able to access sufficient labeled data from target domain for constructing auxiliary tasks. Meanwhile, obtaining the transfer performance of auxiliary tasks needs retraining the source model, which incurs additional computation costs. As a result, the reliance on auxiliary tasks makes OTCE relatively inefficient and less applicable in general scenarios. 

The statistic of the learned parameters $\lambda_1, \lambda_2, b$ (shown in Fig. \ref{fig: coefficients analysis}) reveals that $\frac{|\lambda_2|}{|\lambda_1|}$ among different transfer configurations varied irregularly, suggesting that the importance of domain difference and task difference varies for different cross-domain transfer learning settings. It is therefore incapable of using the pre-defined coefficients for computing OTCE scores. Additionally, we notice that the task difference $W_T$ plays a more important role ($\frac{|\lambda_2|}{|\lambda_1|} > 1$) in evaluating transferability. Therefore, our proposed auxiliary-free transferability metrics mainly utilize the task difference for describing transferability.
 
}

\section{Auxiliary-free Transferability Metrics}
\label{sec: two auxiliary metrics}

Our proposed auxiliary-free transferability metrics \textbf{F-OTCE (Fast OTCE)} and \textbf{JC-OTCE (Joint Correspondence OTCE)} can be viewed as the efficient versions of the auxiliary-based OTCE metric, which only consider the Negative Conditional Entropy to describe transferability, as depicted in Fig. \ref{fig:otce framework comparison}. Although we do not explicitly evaluate the domain difference, the estimated probabilistic coupling between the source and target data implicitly captures the domain difference to some extent in this unified framework. 

\revise{Specifically, F-OTCE achieves higher efficiency, while JC-OTCE performs better in terms of accuracy across diverse scenarios. The main difference between the two metrics is that the ground cost of  JC-OTCE considers both sample distance and label distance when calculating the optimal coupling between the source and target data, which approximates computing ground cost in the joint space $\mathcal{X} \times \mathcal{Y}$, resulting in more precise data correspondences.}

\subsection{ F-OTCE Metric}
\label{sec: otce-s metric}

Formally, we first use the source feature extractor $\theta_s$ to embed the source and target input instances as latent features, denoted as $\hat{x}^i_s = \theta_s(x^i_s)$ and $\hat{x}^i_t = \theta_s(x^i_t)$ respectively. Then the computation process contains two steps as described below. \par

\textbf{Step1: Compute optimal coupling.} First, for the F-OTCE metric, we define the ground cost between samples as:
\begin{equation}
	c_1(\hat{x}^i_s, \hat{x}^j_t) \triangleq \| \hat{x}^i_s - \hat{x}^j_t \| ^2_2,
	\label{eq:cost for otce-s}
\end{equation}
so the OT problem with the entropic regularization~\cite{cuturi2013sinkhorn} can be defined as:
\begin{equation}
		OT(X_s, X_t)  \triangleq  \mathop{\min}\limits_{\pi \in \mathcal{P}(X_s, X_t)}  \sum_{i,j=1}^{m,n}  c_1(\hat{x}^i_s,\hat{x}^j_t)\pi_{ij} - \lambda H(\pi), 
	\label{eq: OT for OTCE-S}
\end{equation}
where $\pi$ is the coupling matrix of size $m\times n$, and $H(\pi)=-\sum_{i=1}^m \sum_{j=1}^n \pi_{ij}\log\pi_{ij}$ is the entropic regularizer with $\lambda=0.1$. The OT problem above can be solved efficiently by the Sinkhorn algorithm~\cite{cuturi2013sinkhorn} to produce an optimal coupling matrix $\pi^*$. \par

From a probabilistic point of view, the coupling matrix $\pi^*$ is a non-parametric estimation of the joint probability distribution of the source and target latent features $P(X_s,X_t)$. We model the relationship between the source and the target data according to the following simple Markov random field: $Y_s-X_s-X_t-Y_t$, where label random variables $Y_s$ and $Y_t$ are only dependent on $X_s$ and $X_t$, respectively, i.e., $P(Y_s,Y_t|X_s,X_t) = P(Y_s|X_s)P(Y_t|X_t)$.
Furthermore, we can derive the empirical joint probability distribution of the source and target labels,
\begin{equation}
	P(Y_s,Y_t) = \mathbb{E}_{X_s,X_t}[P(Y_s|X_s)P(Y_t|X_t)].
	\label{eq: joint label distribution}
\end{equation}
This joint probability distribution can reveal the transfer performance since the goodness of class-to-class matching intuitively reveals the hardness of transfer.\par

\textbf{Step2: Compute negative conditional entropy.} We are inspired by Tran \textit{et al.}~\cite{NCE} who use Conditional Entropy (CE) $H(Y_t|Y_s)$ to describe class-to-class matching quality over the same input instances. They have shown that the empirical transferability is lower bounded by the negative conditional entropy, 
\begin{equation}
	\widetilde{\mathrm{Trf}}(S \rightarrow T) \ge l_S(\theta_s, h_s) - H(Y_t|Y_s),
	\label{eq: transferability inequality}
\end{equation} 
where the training log-likelihood $\widetilde{\mathrm{Trf}}(S \rightarrow T)  = l_T(\theta_s, h_t) =  \frac{1}{n}\sum^n_{i=1}\log P(y^i_t|x^i_t; \theta_s,h_t)$ is an approximation of the empirical transferability when the retrained model is not overfitted. And $l_S(\theta_s, h_s)$ is a constant, so the empirical transferability can be attributed to the conditional entropy.\par

We consider it as a reasonable metric to evaluate transferability under the cross-domain cross-task transfer setting once we learn the soft correspondence $\pi^*$ between source and target features via optimal transport. We can also compute the empirical joint probability distribution of the source and target labels, and the marginal probability distribution of the source label, denoted as: 
\begin{equation}
	\hat{P}(y_s,y_t) = \sum_{i,j: y^i_s=y_s, y^j_t=y_t} \pi_{ij}^*,
	\label{eq: joint distribution}
\end{equation}

\begin{equation}
	\hat{P}(y_s) = \sum_{y_t \in \mathcal{Y}_t} \hat{P}(y_s,y_t).
	\label{eq: marginal distribution}
\end{equation}
Then we can compute the negative conditional entropy as the F-OTCE score,  
\begin{equation}
	\begin{aligned}
		\text{F-OTCE} & = - H_{\pi^*}(Y_t|Y_s)\\
		& = \sum_{y_t \in \mathcal{Y}_t} \sum_{y_s \in \mathcal{Y}_s} \hat{P}(y_s,y_t)\log \frac{\hat{P}(y_s,y_t)}{\hat{P}(y_s)}.
	\end{aligned}
	\label{eq: otce-s}
\end{equation}\par 

Compared to the auxiliary-based OTCE, we directly use the negative conditional entropy to characterize transferability, which avoids the cumbersome parameter fitting process on auxiliary tasks, resulting in a drastic efficiency improvement.

\begin{figure}[t]
	\centering
	\includegraphics[width=1.0\linewidth]{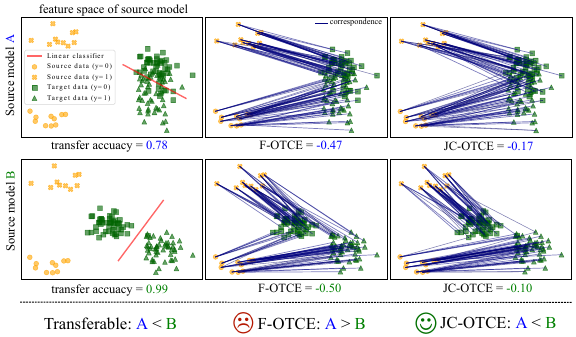}
	\caption{A toy example shows that the F-OTCE metric fails to distinguish the more transferable source model, while the JC-OTCE predicts correctly by involving the label distance in computing the correspondences.}
	\label{fig: toy example}
\end{figure}

\subsection{ JC-OTCE Metric}
\label{sec: jc-otce metric}

F-OTCE is an efficient transferability metric in practical scenarios, but its accuracy can be further improved. Take a toy example shown in Fig. \ref{fig: toy example} for illustration, where the F-OTCE metric fails to distinguish the more transferable source model. This observation suggests that computing data correspondences solely based on sample distance (in space $\mathcal{X}$) may not always accurately capture the class-to-class matching quality (or the label uncertainty of the target task) as expected. \revise{Therefore, to further improve the accuracy of F-OTCE, we propose the JC-OTCE metric which involves the additional label distance in computing the joint correspondences between data in the joint space $\mathcal{X} \times \mathcal{Y}$.} \par

Formally, we first define the data instances of the source and target tasks as $z_s=(\hat{x}_s,y_s)$ and $z_t=(\hat{x}_t,y_t)$ respectively, where $z_s \in \mathcal{Z}_s = \mathcal{X} \times \mathcal{Y}_s$ and $z_t \in \mathcal{Z}_t = \mathcal{X} \times \mathcal{Y}_t$. And we define the $\alpha_y \triangleq P(X | Y=y)$, which can be estimated from a collection of finite samples with label $y$. Inspired by recent work~\cite{OTDD}, we compute the label distance as the Wasserstein distance $Wass(\alpha_{y_s}, \alpha_{y_t})$. Then the ground cost for JC-OTCE can be defined as:

\begin{equation}
	c_2(z^i_s, z^j_t) \triangleq \gamma \| \hat{x}^i_s - \hat{x}^j_t \| ^2_2 + (1-\gamma)Wass(\alpha_{y^i_s}, \alpha_{y^j_t}),
	\label{eq: cost of JC-OTCE}
\end{equation}
where $\gamma \in [0,1]$ is a weighting coefficient to combine the sample distance and the label distance, and here we let $\gamma=0.5$. More discussion about $\gamma$ is described in Section \ref{sec: effect of gamma}. Similarly, the OT problem for $Z_s$ and $Z_t$ is defined as:
\begin{equation}
	OT(Z_s, Z_t)  \triangleq  \mathop{\min}\limits_{\pi \in \mathcal{P}(Z_s, Z_t)}  \sum_{i,j=1}^{m,n}  c_2(z^i_s,z^j_t)\pi_{ij} - \lambda H(\pi).
	\label{eq: OT for JC-OTCE}
\end{equation}

By solving this OT problem, we also obtain the optimal coupling matrix $\pi ^*$. Then following the \textbf{Step2} described in previous Section \ref{sec: otce-s metric} (see Equation (\ref{eq: joint distribution}), (\ref{eq: marginal distribution})), the JC-OTCE score is computed as the negative conditional entropy as well.
\begin{equation}
	\begin{aligned}
		\text{JC-OTCE} & = - H_{\pi^*}(Y_t|Y_s)\\
		& = \sum_{y_t \in \mathcal{Y}_t} \sum_{y_s \in \mathcal{Y}_s} \hat{P}(y_s,y_t)\log \frac{\hat{P}(y_s,y_t)}{\hat{P}(y_s)}.
	\end{aligned}
	\label{eq: otce-s}
\end{equation}

\section{Transferability-guided Transfer Learning}
\label{sec: otce-based finetune}
 
\revise{In this section, we present two examples of utilizing our transferability metric to boost the performance of downstream transfer learning tasks, including model finetuning and domain generalization. The differences between these two transfer learning tasks are described in Table \ref{tab: method difference}. 
	
To facilitate the training process, we adopt the F-OTCE metric as the optimization objective since using the JC-OTCE metric needs solving multiple OT problems to compute pair-wise label distances, which incurs significant computational costs. Additionally, due to GPU memory constraints, we typically perform mini-batch training which only loads a subset of the dataset in the current training iteration, while computing label distance requires loading the entire dataset.}

\begin{table}[t]
	\centering
\begin{minipage}{1.0\linewidth}

\centering


\caption{\revise{Differences between model finetuning and domain generalization.}}
\label{tab: method difference}

\begin{tabular}{c|c|c}
\toprule
& Model finetuning & Domain generalization \\ 

\midrule 

Source data & Single & Multiple \\

Target task & Known & Unknown \\

\multirow{2} {*}{Goal} & Achieving higher accuracy  & Learning generalizable \\
& on the target task & feature representations\\

\bottomrule

\end{tabular}

\end{minipage}
\end{table}

\subsection{OTCE-based Model Finetuning}

\revise{The vanilla finetune algorithm follows the ``pretraining + finetuning'' pipeline that is commonly used in transfer learning. However, this scheme does not consider the relatedness between the source and target tasks. To address this issue, our proposed OTCE-based finetune algorithm introduces an intermediate step into the conventional pipeline, i.e., maximize the transferability of transferring from the source task to the target task, resulting in a ``pretraining + adaptation (maximizing transferability) + finetuning'' framework. The moderate optimization during the adaptation step utilizes the task relationship characterized by our F-OTCE score to enable the source feature representation to become more transferable to the target task. This facilitates easier learning of the head classifier during the finetuning step and ultimately leads to higher transfer accuracy.}

\begin{figure}[t]
\centering

\includegraphics[width=0.8\linewidth]{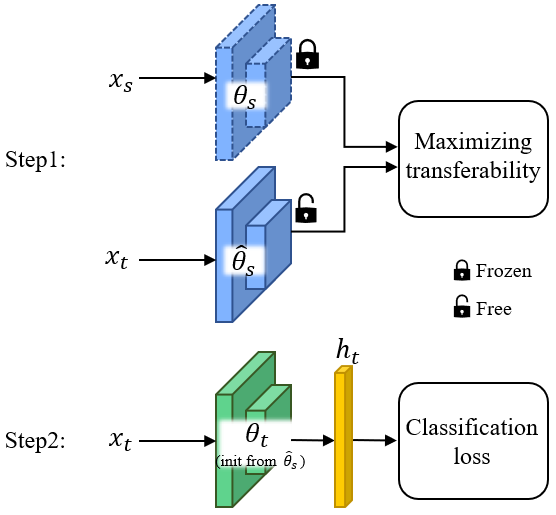}
\caption{The pipeline of our OTCE-based finetune algorithm.}
\label{fig: otce finetune}
\end{figure}

Suppose we have obtained the pretrained model on the source task, the OTCE-based finetune algorithm is a two-step framework, as depicted in Fig. \ref{fig: otce finetune} and Algorithm \ref{alg: OTCE-based finetune}. First, we optimize the source feature extractor $\hat{\theta}_s$ by minimizing the conditional entropy within one epoch. Formally,
\begin{equation}
	\begin{aligned}
		\hat{\theta}^*_s & =\mathop{\arg\min}_{\hat{\theta}_s} \ H_{\pi^*}(Y_t|Y_s) \\
		& =  -  \mathop{\arg\min}_{\hat{\theta}_s} \sum_{y_t \in \mathcal{Y}_t} \sum_{y_s \in \mathcal{Y}_s} \hat{P}(y_s,y_t)\log \frac{\hat{P}(y_s,y_t)}{\hat{P}(y_s)},
	\end{aligned} 
	\label{eq: task loss}
\end{equation}
where the $\pi^*$ is the optimal coupling matrix computed from Equation (\ref{eq: OT for OTCE-S}). Joint label distribution $\hat{P}(y_s,y_t)$ and marginal $\hat{P}(y_s)$ are computed from Equation (\ref{eq: joint distribution}), (\ref{eq: marginal distribution}). The computation of solving the OT problem with entropic regularizer~\cite{cuturi2013sinkhorn} (Equation (\ref{eq: OT for OTCE-S})) is differentiable~\cite{genevay2018learning} since the iterations form a sequence of linear operations, so it can be implemented on the PyTorch framework as a specialized layer\footnote{https://github.com/dfdazac/wassdistance} of the neural network. After that, we initialize the target feature extractor $\theta_t$ from the optimized source weights $\hat{\theta}^*_s$, and then retrain the target model $(\theta_t, h_t)$ on the target training data using the cross-entropy loss function, 
\begin{equation}
	\theta^*_t, h^*_t = \mathop{\arg\max}_{\theta_t, h_t}\   \sum_{i=1}^{m} \sum_{l=1}^{k} \mathbf{1}\{y^i_t=l\} \log \frac{\exp(h^l_t(\theta_t(x^i_t)))}{\sum_{j=1}^k\exp(h^j_t(\theta_t(x^i_t)))}, 
	\label{eq: classification loss}
\end{equation}  
where $m$ represents the number of target training samples, and $k$ is the number of the categories of the target task.\par

Note that we do not make it a one-step framework, i.e., simultaneously maximize the transferability and minimize the classification loss. Because optimizing two objectives simultaneously may cause gradient conflicts in mini-batch training, which will deteriorate the final classification performance.

\begin{algorithm}[t]
	\caption{OTCE-based finetune}
	\label{alg: OTCE-based finetune}
	\begin{algorithmic}[1]
		{\footnotesize
			\REQUIRE 
			source dataset $D_s = \{(x^i_s, y^i_s)\}_{i=1}^{m}$\\
			\quad \quad \;target dataset $D_t = \{(x^i_t, y^i_t)\}_{i=1}^{n}$ \\
			\quad \quad \;source feature extractor $\theta_s$ \\
			
			\vspace{5pt}
			
			\STATE{Initialize} $\hat{\theta}_s = \theta_s$
			\WHILE{\text{sampling mini-batches within one epoch}}
			
			\STATE{Generate mini-batch} $B_s = \{(\theta_s(x^i_s),y^i_s)\}^M_i$
			\STATE{Generate mini-batch} $B_t = \{(\hat{\theta}_s(x^i_t),y^i_t)\}^N_i$
			
			\STATE{Update} $\hat{\theta}_s$ via maximizing $\text{F-OTCE} (B_s, B_t)$
			
			\ENDWHILE
			
			\STATE{Initialize} $\theta_t = \hat{\theta}_s$
			\STATE{Randomly initialize} $h_t$
			\WHILE{$\theta_t, h_t$ \text{not converge}}
			
			\STATE{Update} $\theta_t, h_t$ using equation (\ref{eq: classification loss})
			\ENDWHILE

		}
	\end{algorithmic}
\end{algorithm}

\subsection{\revise{OTCE-based Domain Generalization}}

\revise{In contrast to the model finetuning task, the domain generalization (DG) task aims to learn the generalizable feature representation exhibiting domain-irrelevant and task-irrelevant characteristics from multiple training domains. Therefore, the learned model can also achieve high classification accuracy when transferred to the unseen tasks from unseen domains. We integrate our F-OTCE metric into a state-of-the-art domain generalization method URL~\cite{li2022universal, li2021universal} as a loss function to illustrate its effectiveness in boosting the DG algorithm.}

\revise{ More specifically, URL learns a universal model via distilling the common knowledge from multiple pretrained domain-specific models corresponding to each training domain. The universal model is required to achieve high classification accuracy in all training domains as well. Once the universal model is obtained, we can use it to extract feature representations for unseen few-shot classification tasks and make predictions via the nearest neighbor classifier (NCC).}

\begin{figure}[t]
	\centering
	\includegraphics[width=0.9\linewidth]{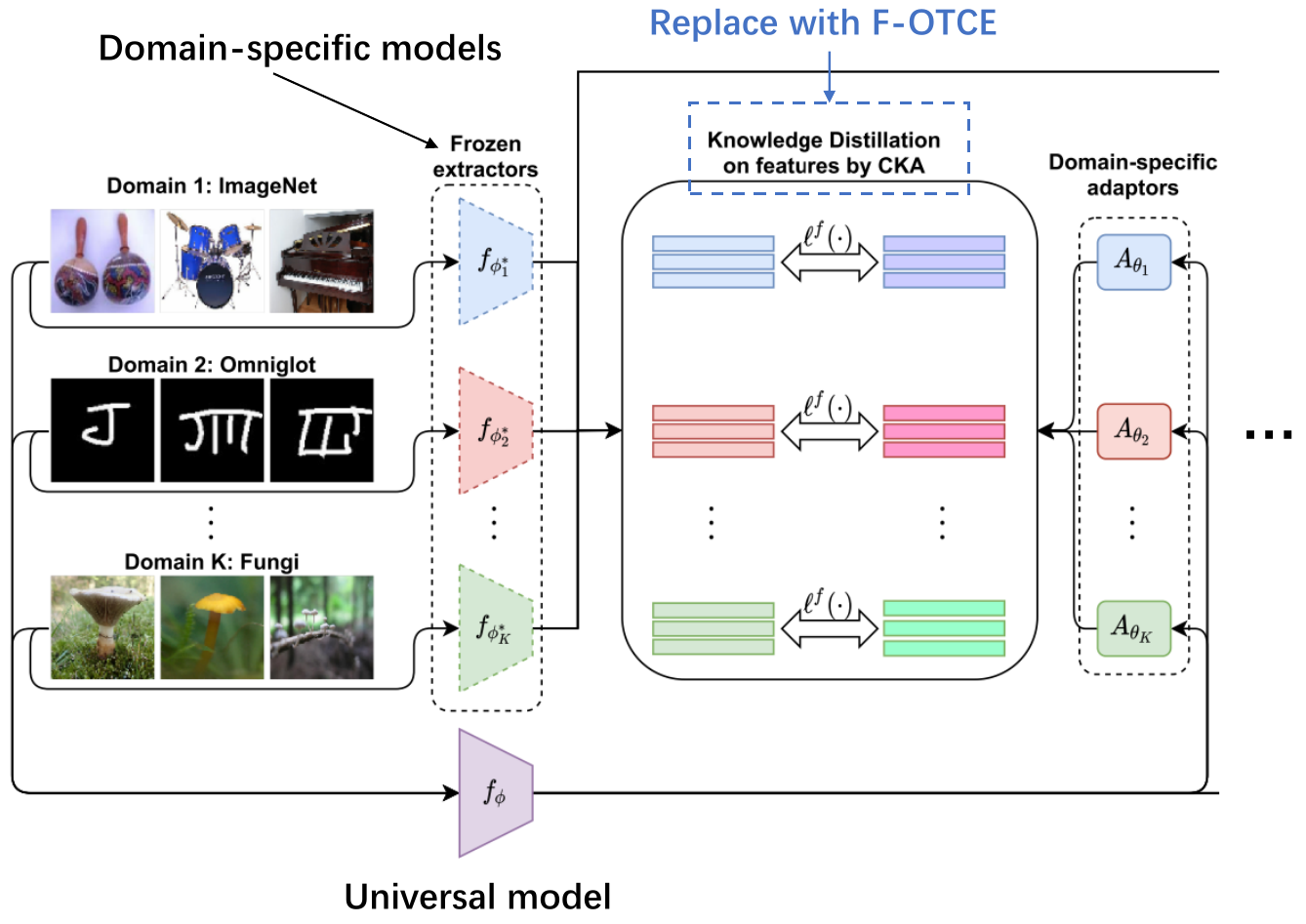}
	\caption{\revise{Partial illustration of the URL framework~\cite{li2021universal}. We replace the CKA similarity with our F-OTCE metric.}}
	\label{fig: URL}
	\vspace{-0.5cm}
\end{figure}

\revise{In our opinion, the process of distilling knowledge from domain-specific models can be interpreted as maximizing the transferability between the domain-specific models and the universal model. Therefore, we propose to replace the knowledge distillation objective Centered Kernel Alignment (CKA)~\cite{kornblith2019similarity} similarity used in URL with our F-OTCE metric, as illustrated in Fig. \ref{fig: URL}. Unlike CKA which solely focuses on minimizing feature differences, F-OTCE considers a wider range of task-specific information to minimize the label uncertainty of the universal model. We follow the default configuration of the URL algorithm. Please refer to \cite{li2022universal, li2021universal} and the official codebase\footnote{https://github.com/VICO-UoE/URL} for more details about the URL algorithm.}

\subsection{Few-shot Classification Task Definition}

We evaluate the effectiveness of our algorithms based on their transfer accuracy on few-shot classification tasks across domains. A few-shot classification task known as \textit{C-way-K-shot} means that the support (training) set $S =\{(x^i, y^i) \}_{i=1}^{k\times C}$ contains $k$ labeled instances from each of the $C$ categories. The query set $Q =\{(x^i, y^i) \}_{i=1}^{q\times C}$ contains $q$ samples per category and serves as the testing set to evaluate the classification accuracy of the model finetuned on the support set.
\section{Experiments}
\label{sec: experiment}

\revise{In this section, we begin by conducting quantitative evaluations of our proposed transferability metrics under various cross-domain cross-task transfer settings. We also explore their applications in source model selection and multi-source feature fusion, as well as provide further analyses on computation efficiency, memory consumption and hyperparameters. Additionally, we conduct extensive evaluations of our proposed transferability-guided transfer learning methods including the OTCE-based finetune algorithm and the OTCE-based URL algorithm.}

\subsection{Evaluation on Transferability Estimation}

\begin{figure}[t]
\centering
\includegraphics[width=0.9\linewidth]{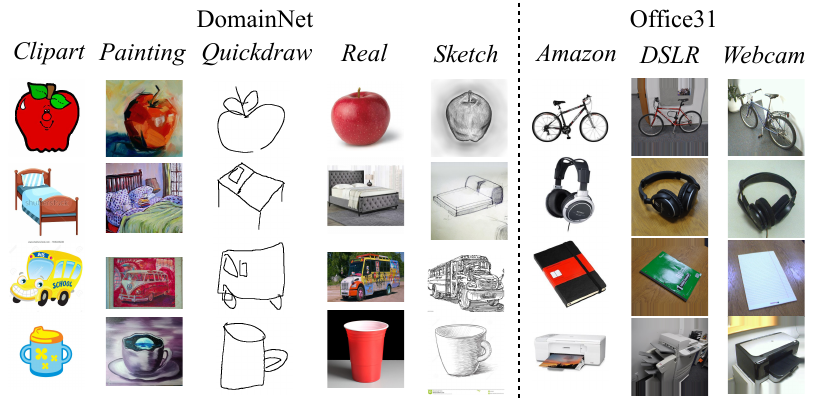}
\caption{Examples from the cross-domain datasets DomainNet and Office31, where images from different domains exhibit different image styles or are captured by different devices.}
\label{fig: vis cls dataset}
\end{figure}

\textbf{Datasets.} Our experiments are conducted on the data from the largest-to-date cross-domain dataset DomainNet~\cite{DomainNet} and popular Office31~\cite{Office31} dataset. The DomainNet dataset contains 345-category images in five domains (image styles), i.e., Clipart (C), Painting (P), Quickdraw (Q), Real (R), and Sketch (S), and the Office31 contains 31-category images in three domains including Amazon (A), DSLR (D) and Webcam (W). Data examples are shown in Fig. \ref{fig: vis cls dataset}.\par 

\textbf{Evaluation criteria.} To quantitatively evaluate the effectiveness of transferability metrics, we adopt the commonly-used Spearman's rank correlation coefficient (Spearman's $\rho$ coefficient) and the Kendall rank correlation coefficient (Kendall's $\tau$ coefficient)~\cite{kendall} to assess the correlation between the transfer accuracy and predicted transferability scores. Specifically, the Spearman's $\rho$ coefficient is defined as: 

\begin{equation}
	\rho = 1- \frac{6\sum d_i^2}{n(n^2-1)}, 
\end{equation}
where $d_i = R(\text{Acc}_i) - R(\text{Trf}_i)$  is the difference between the rankings of transfer accuracy $\text{Acc}_i$ and transferability score $\text{Trf}_i$ for the $i$th source-target task pair, and $n$ represents the total number of task  pairs. \par 

The Kendall's $\tau$ coefficient in our experiments is defined as:
\begin{equation}
	\tau = \frac{2}{n(n-1)}\sum_{i<j}\text{sgn}(\text{Acc}_i - \text{Acc}_j)\text{sgn}(\text{Trf}_i-\text{Trf}_j).
\end{equation}
The Kendall's $\tau$ coefficient computes the number of concordant pairs minus the number of discordant pairs divided by the number of total pairs. A higher rank correlation indicates the more accurate transferability estimation result.\par

\begin{table*}[t]
	\centering
\begin{minipage}{1.0\linewidth}

\centering
\setlength\tabcolsep{4pt} 

\footnotesize

\caption{\minorrevise{Quantitative comparisons evaluated by Spearman's $\rho$ coefficient and Kendall's $\tau$ coefficient between transferability metrics and transfer accuracy under different cross-domain cross-task transfer settings for image classification tasks. Our proposed JC-OTCE and F-OTCE metrics consistently outperform state-of-the-art auxiliary-free metrics. Meanwhile, the JC-OTCE achieves comparable performance to the auxiliary-based OTCE.}}
\label{tab: transferablity score for classification}

\begin{tabular}{ccc|c|cccccc}
	\toprule
	
	\multirow{4} {*}{Setting} &  & & \multicolumn{7}{c}{Spearman / Kendall correlation coefficient} \\
	\cmidrule{4-10}
	& Source& Target & \textcolor{Grey}{Auxiliary-based} & \multicolumn{6}{c}{Auxiliary-free} \\
	\cmidrule{4-10}
	& domain & domain& \textcolor{Grey}{OTCE~\cite{OTCE}} & JC-OTCE & F-OTCE & LEEP~\cite{LEEP} &NCE~\cite{NCE} & H-score~\cite{bao2019information}& LogME~\cite{LogME}\\
	
	\midrule
	
	&  C & P,Q,R,S  & \textcolor{Grey}{0.976 / 0.861}&  \underline{0.965} / \underline{0.836} & \textbf{0.966} / \textbf{0.839} &  0.932 / 0.779 & 0.825 / 0.670& 0.920 / 0.748&0.867 / 0.667\\
	
	& P & C,Q,R,S& \textcolor{Grey}{0.977 / 0.868} & \textbf{0.966} / \textbf{0.837} &\underline{0.960} / \underline{0.822} & 0.906 / 0.743& 0.849 / 0.686 & 0.937 / 0.777 & 0.929 / 0.761 \\
	
	Standard& Q & C,P,R,S& \textcolor{Grey}{0.961 / 0.826}  & \underline{0.962} / \textbf{0.833} &\textbf{0.963} / \underline{0.832}& 0.953 / 0.810 & 0.943 / 0.793 & 0.942 / 0.784 & 0.912 / 0.744 \\
	
	(Retrain head) & R& C,P,Q,S & \textcolor{Grey}{0.975 / 0.863}& \textbf{0.965} / \textbf{0.836} & \underline{0.951} / \underline{0.808} & 0.910 / 0.747 & 0.872 / 0.707 & 0.942 / 0.786 & 0.855 / 0.670\\
	
	& S & C,P,Q,R & \textcolor{Grey}{0.969 / 0.842}& \underline{0.965} / \underline{0.834} & \textbf{0.967} / \textbf{0.839}& \underline{0.965} / \underline{0.834} & 0.962 / 0.830& 0.950 / 0.802 &0.908 / 0.733\\
	
	\cmidrule{2-10}

	&  C & P,Q,R,S  & \textcolor{Grey}{0.932 / 0.766} & \textbf{0.900} / \textbf{0.713} & 0.884 / 0.689 & 0.814 / 0.618 & 0.664 / 0.517 & 0.889 / \underline{0.704} & \underline{0.890} / 0.695 \\

	& P & C,Q,R,S& \textcolor{Grey}{0.803 / 0.612} & 0.874 / \underline{0.698} & \textbf{0.880} / \underline{0.698} & 0.850 / 0.655 & 0.797 / 0.613 & \underline{0.876} / \textbf{0.716} & 0.848 / 0.664 \\
	
	\revise{Standard} & Q & C,P,R,S& \textcolor{Grey}{0.896 / 0.719} & \textbf{0.906} / \textbf{0.732} & \underline{0.895} / \underline{0.719} & 0.880 / 0.696 & 0.874 / 0.684 & 0.873 / 0.686 & 0.891 / 0.699 \\
	
	\revise{(Finetune)}& R& C,P,Q,S & \textcolor{Grey}{0.912 / 0.732} & \textbf{0.905} / \underline{0.725} & 0.882 / 0.689 & 0.821 / 0.616 & 0.770 / 0.571 & \underline{0.902} / \textbf{0.727} & 0.876 / 0.681 \\
	
	& S & C,P,Q,R & \textcolor{Grey}{0.923 / 0.752} & \textbf{0.932} / \textbf{0.767} & \underline{0.929} / 0.763 & 0.927 / \underline{0.766} & 0.925 / 0.757 & 0.915 / 0.747 & 0.894 / 0.706\\

	\cmidrule{2-10}
	
	& & Average & \textcolor{Grey}{0.932 / 0.784} & \textbf{0.934} / \textbf{0.782} & \underline{0.928} / \underline{0.770} & 0.896 / 0.727 & 0.849 / 0.682 & 0.915 / 0.748 & 0.887 / 0.702 \\
	
	\midrule
	\midrule
	
	 & C & P,Q,R,S & \textcolor{Grey}{0.926 / 0.756} & \textbf{0.926} / \textbf{0.757} &\underline{0.909} / \underline{0.729} & 0.836 / 0.640 & 0.745 / 0.576 & 0.762 / 0.567 & 0.731 / 0.524\\
	
	& P & C,Q,R,S& \textcolor{Grey}{0.931 / 0.772} & \textbf{0.928} / \textbf{0.769}& \underline{0.886} / \underline{0.701} & 0.803 / 0.618 & 0.746 / 0.575 & 0.811 / 0.608 & 0.849 / 0.649\\
	
	Few-shot& Q & C,P,R,S & \textcolor{Grey}{0.821 / 0.631} & \underline{0.856} / \underline{0.673} & 0.829 / 0.636 & 0.798 / 0.602 & 0.782 / 0.584 & 0.813 / 0.614 & \textbf{0.866} / \textbf{0.682}\\
	
	(Retrain head)& R& C,P,Q,S& \textcolor{Grey}{0.929 / 0.769} & \textbf{0.897} / \textbf{0.724} & \underline{0.853} / \underline{0.666}& 0.770 / 0.589& 0.728 / 0.559 & 0.845 / 0.652 & 0.774 / 0.574\\
	
	& S & C,P,Q,R & \textcolor{Grey}{0.914 / 0.742} & \textbf{0.902} / \textbf{0.725} & \underline{0.895} / \underline{0.710} & 0.872 / 0.680 & 0.872 / 0.679 & 0.838 / 0.645 & 0.867 / 0.684\\

	\cmidrule{2-10} 
	& & Average & \textcolor{Grey}{0.905 / 0.734} & \textbf{0.902} / \textbf{0.729} & \underline{0.875} / \underline{0.689} & 0.815 / 0.625 & 0.775 / 0.595 & 0.814 / 0.618 & 0.818 / 0.623 \\
	
	\midrule
	\midrule

	&   C & P,Q,R,S & \textcolor{Grey}{0.701 / 0.500} & \textbf{0.695} / \textbf{0.498} & \underline{0.687} / \underline{0.487} & 0.685 / 0.486& 0.666 / 0.472 &-0.438 / -0.290 & -0.222 / -0.151\\
Fixed category 	& P & C,Q,R,S & \textcolor{Grey}{0.670 / 0.485} & \textbf{0.665} / \textbf{0.479} & \underline{0.631} / \underline{0.448} & 0.630 / 0.446 &0.612 / 0.430&-0.529 / -0.371&-0.043 / -0.039\\
	
	 size & Q & C,P,R,S & \textcolor{Grey}{0.341 / 0.225} & \textbf{0.381} / \textbf{0.261} & \underline{0.316} / \underline{0.211}&0.210 / 0.136& 0.291 / 0.191 & -0.256 / -0.172&0.066 / 0.037\\
	
	(Retrain head)&  R& C,P,Q,S & \textcolor{Grey}{0.637 / 0.455} & \textbf{0.695} / \textbf{0.498} & \underline{0.598} / \underline{0.415} & 0.587 / 0.407 &0.586 / 0.406&-0.094 / -0.063&-0.382 / -0.252\\
	
	&  S & C,P,Q,R & \textcolor{Grey}{0.428 / 0.292} & \textbf{0.497} / \textbf{0.343} & \underline{0.436} / \underline{0.299} &0.404 / 0.277& 0.432 / 0.298 & -0.247 / -0.164&0.027 / 0.006\\
	
	\cmidrule{2-10} 
	& & Average & \textcolor{Grey}{0.555 / 0.391} & \textbf{0.587} / \textbf{0.416} & \underline{0.534} / \underline{0.372} &0.503 / 0.350 & 0.517 / 0.359 &-0.313 / -0.212&-0.111 / -0.080\\
	
	\midrule
\midrule

\minorrevise{Imbalanced}	&   A & D & \textcolor{Grey}{- / -} & \textbf{0.844} / \textbf{0.646} & \underline{0.829} / \underline{0.627} & 0.822 / 0.616 & 0.801 / 0.589 & 0.674 / 0.476 & 0.785 / 0.593 \\
(Retrain head)	&   A & W &\textcolor{Grey}{- / -} & 0.847 / 0.651 & 0.850 / 0.653 & \textbf{0.862} / \textbf{0.665} & \underline{0.859} / \underline{0.663} & 0.657 / 0.489 & 0.787 / 0.590 \\
	
	\cmidrule{2-10} 
	
\revise{Balanced}	&   A & D & \textcolor{Grey}{- / -} & \underline{0.822} / \textbf{0.627} & \textbf{0.824} / \underline{0.625} & 0.796 / 0.592 & 0.783 / 0.572 & 0.574 / 0.393 & 0.747 / 0.536 \\
(Retrain head)	&   A & W & \textcolor{Grey}{- / -} & \textbf{0.879} / \textbf{0.686} & 0.871 / 0.673 & \underline{0.872} / \underline{0.674} & 0.856 / 0.656 & 0.669 / 0.477 & 0.797 / 0.604 \\
	\cmidrule{2-10}
	& & Average &\textcolor{Grey}{- / -} & \textbf{0.848} / \textbf{0.653} & \underline{0.844} / \underline{0.645} & 0.838 / 0.637 & 0.825 / 0.620 & 0.644 / 0.459 & 0.779 / 0.581\\
	
	\bottomrule
	
\end{tabular}
\newline
\footnotesize{\textbf{Bold} denotes the best result, and \underline{underline} denotes the \engordnumber{2} best result.}
\end{minipage}

\vspace{-0.3cm}
\end{table*}

\begin{figure*}[t]
\centering
\def\svgwidth{1.0\linewidth}
	\executeiffilenewer{images/transferability_classification.svg}{images/transferability_classification.pdf}%
	{inkscape -z -D --file=images/transferability_classification.svg %
		--export-pdf=images/transferability_classification.pdf --export-latex}%
	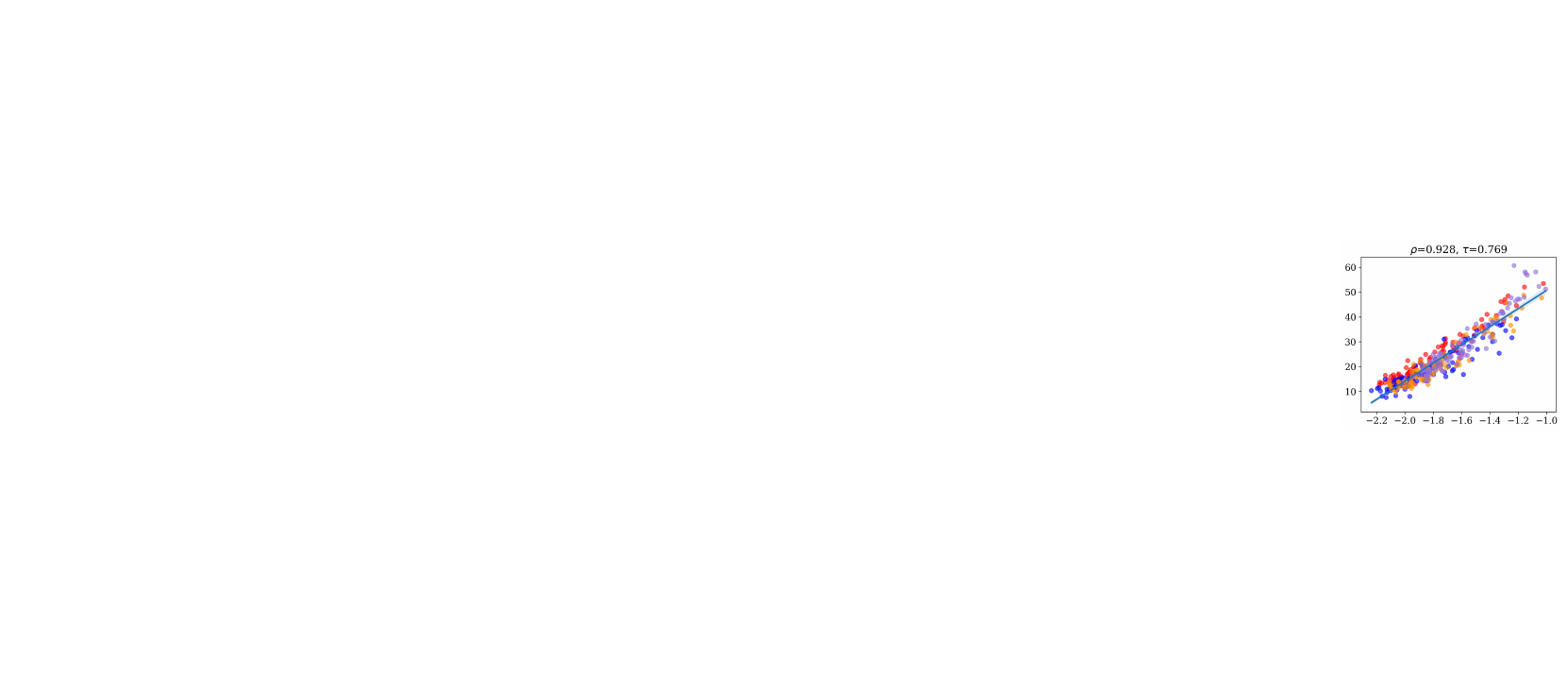%

\caption{Visualization of the correlation between the transfer accuracy and transferability metrics, where the vertical axis denotes the transfer accuracy and the horizontal axis represents the transferability scores. Points in the figure represent different target tasks. Our JC-OTCE and F-OTCE metrics show significantly better correlation (more compact) with the transfer accuracy compared to state-of-the-art auxiliary-free metrics, especially under the challenging \textit{fixed category size setting}. Meanwhile, the JC-OTCE metric performs comparably well to the auxiliary-based OTCE metric.}
\label{fig: transferability classification}
\end{figure*}

\textbf{Transfer settings.} In DomainNet dataset, we successively take each domain as the source domain and use the rest as target domains. For each target domain, we generate 100 target tasks by randomly sampling images in different categories. Then we transfer the source models (ResNet-18~\cite{he2016deep}) pretrained on all source domain data to each target task to obtain the ground-truth transfer accuracy. To investigate the performance of transferability metrics under various transfer configurations, three different transfer settings are considered, i.e., the \textit{standard setting}, the \textit{few-shot setting}, and the \textit{fixed category size} setting.\par 

\begin{itemize}
	\item \textit{Standard setting.} We keep all the training samples of the target task for retraining the source model. Meanwhile, the number of categories of target tasks ranges from 10 to 100. Thus we totally conduct $5\times 4 \times 100 = 2000$ cross-domain cross-task transfer tests.
	
	\item \textit{Few-shot setting.} As transfer learning is commonly used in scenario where only a few labeled data are provided, it is worth evaluating the accuracy of transferability metrics on few-shot cases. The only difference with the \textit{standard setting} is that we limit the target tasks to have only 10 training samples per category.

	\item \textit{Fixed category size setting.} As studied in \cite{OTCE}, the intrinsic complexity of the target task, e.g. category size (number of categories), also affects the transfer accuracy. Usually, a larger category size makes the target task more difficult to learn from limited data. As a result, in the previous two settings, the intrinsic complexity of target tasks with different category sizes may overshadow the more subtle variations in the relatedness with the source task. To investigate whether the transferability metrics are capable of capturing those subtle variations, we propose a more challenging \textit{fixed category size} setting where all target tasks have the same $category\_ size = 50$. Other configurations are the same as the \textit{standard setting}.\par
	 
\end{itemize}

Moreover, in the Office31 dataset, the DSLR and Webcam domains contain very few samples ($\sim$15 samples per category) and suffer from severe category imbalance. Consequently, we construct two different configurations: \textit{data-imbalanced} and \textit{data-balanced} settings. Both of these two settings are few-shot, but the data-balanced setting permits a maximum of 10 samples per category. Here we only use Amazon as the source domain since the other two domains lack sufficient data to train generalizable source models. \minorrevise{It is worth noting that we use the \textit{average per-class accuracy} instead of the \textit{overall accuracy} for representing the transfer performance under the data-imbalanced setting.}

For all the settings above, we adopt an SGD optimizer with a learning rate of 0.01 to optimize the cross-entropy loss for 100 epochs during the transfer training phase.\par

\textbf{Results.} Quantitative comparisons with state-of-the-art auxiliary-free transferability metrics including LEEP~\cite{LEEP}, NCE~\cite{NCE}, H-score~\cite{bao2019information}, LogME~\cite{LogME} are shown in Table \ref{tab: transferablity score for classification}, and visual comparisons are illustrated in Fig. \ref{fig: transferability classification}. Firstly, we can see that both our JC-OTCE and F-OTCE metrics consistently outperform recent LEEP, NCE, H-score, and LogME metrics on all three transfer settings. In particular, our JC-OTCE metric achieves $(7.3\%, 14.7\%, 4.5\%, 11.4\%)$ and $(16.6\%, 22.5\%, 18.0\%, 17.0\%)$ average gains on Kendall correlation compared to LEEP, NCE, H-score, and LogME respectively under the \textit{standard} setting and the \textit{few-shot} setting. Moreover, the H-score metric and the LogME metric failed under the more challenging \textit{fixed category size} setting, where they showed a negative correlation with the transfer accuracy.\par 

Secondly, the JC-OTCE metric outperforms the F-OTCE metric with an average $5.4\%$ gain on Kendall correlation, which shows that involving the label distance in computing the data correspondences makes the transferability estimation more accurate. Meanwhile, the JC-OTCE metric performs comparably to the original OTCE metric in accuracy, while the former one is evidently more efficient and has fewer restrictions.\par 

Basically, we can conclude that $ \text{OTCE} \approx \text{JC-OTCE} > \text{F-OTCE}$ in accuracy and $ \text{OTCE} < \text{JC-OTCE} < \text{F-OTCE}$ in efficiency. These three metrics can be applied flexibly according to the needs of different practical situations.

\subsection{Efficiency Analysis}

Given $d$-dimensional extracted features of  $m$ source samples and $n$ target samples, 
assuming that $|\mathcal{Y}_s|,|\mathcal{Y}_t|<\min(m,n)$, the computational complexity of F-OTCE is $O(mn\max\{d,k\} )$, where $k$ is the number of Sinkhorn iterations in the OT computation. 
Specifically, the worst-case complexity of computing the cost matrix between source and target samples is $O(mnd)$. Solving the OT problem by Sinkhorn algorithm with $\epsilon$-accuracy has complexity $O(mnk)=O(2mn\|c\|^2_\infty / (\lambda \epsilon))$~\cite{chizat2020faster} ,  where 
$\|c\|_\infty=\sup_{(z_s,z_t)\in \mathcal{Z}^2} c(z_s,z_t)$ is the maximum cost between source and target sample features and $\lambda$ is the weighting coefficient of the entropic regularizer. In practice, we usually set a constant parameter $\lambda$ and a stopping criteria with a maximum iteration. Finally, the conditional entropy computation takes $O(mn)$ time.\par

Compared to F-OTCE, the additional computation of JC-OTCE lies in computing pair-wise label distances, which needs to solve $|\mathcal{Y}_s|\times |\mathcal{Y}_t|$ OT problems between the samples with given labels and finally produce Wasserstein distances. \par  

\minorrevise{The experimental computation time statistics of the empirical transferability and analytical transferability metrics are presented in Table \ref{tab: efficiency analysis}. Specifically, the empirical transferability calculation is performed on a GPU (NVIDIA GTX1080Ti) by retraining the source model on the target training data and subsequently evaluating its transfer accuracy on the testing set. In order to ensure a fair comparison among analytical transferability metrics, we establish a standardized evaluation setting. For each target classification task, we randomly select 1,000 samples for computation on CPU, which follows the same configuration as in our transferability experiments. To accurately reflect the real wall-clock time required for transferability estimation, we have excluded the time spent on I/O operations. We conducted these computations over 10 random tasks and calculated the average time as the final results. The overall computation time consists of two components: feature extraction time and transferability prediction time. Importantly, the computation time remains consistent across different types of datasets once the source model architecture and the number of samples have been determined. The memory requirements of F-OTCE and JC-OTCE are 395MB and 407MB respectively, which can be easily met by a typical personal computer.}

Results show that analytical transferability metrics are $\sim 90\times$ faster than the empirical transferability, without the requirement of GPU. Meanwhile, auxiliary-free metrics perform comparably on efficiency costing $\sim$10s for a pair of tasks. Although the original OTCE metric contains additional auxiliary time since it requires at least three auxiliary tasks with known empirical transferability to determine weighting coefficients, it is still worth using the OTCE metric for more accurate transferability estimation when there are many target tasks under the same cross-domain configuration needing evaluations.

\begin{table}[t]
	\centering
\begin{minipage}{1.0\linewidth}

\centering


\caption{\revise{Computation time statistics under the standard evaluation setting. Auxiliary-free metrics achieve comparable efficiency ($\sim 10$s), which is evidently more efficient than the auxiliary-based OTCE and the empirical transferability.}}
\label{tab: efficiency analysis}

\begin{tabular}{lccc}
\toprule
\multirow{2} {*}{Metric} & Auxiliary & Wall-clock & Correlation \\
& time & time & (Spearman) \\
\midrule 

 Empirical  & \multirow{2} {*}{-} & \multirow{2}{*}{858s (14.3min)} & \multirow{2}{*}{1.000}\\
 transferability &  & \\
 \midrule
LEEP~\cite{LEEP} & - & 8.97s & 0.896\\
NCE~\cite{NCE} & - & 8.92s & 0.849\\
H-score~\cite{bao2019information} & - & 9.02s & 0.915 \\
LogME~\cite{LogME} & - & 9.23s & 0.887\\
OTCE~\cite{OTCE} & 858s $\times$ 3 & 9.32s & 0.932\\
\midrule 
F-OTCE & - & 9.32s & 0.928\\
JC-OTCE & - & 10.78s & 0.934 \\

\bottomrule

\end{tabular}

\end{minipage}
\end{table}

\subsection{Effect of Parameter $\gamma$}
\label{sec: effect of gamma}

\begin{figure}[t]
    \centering
    \includegraphics[width=0.85\linewidth]{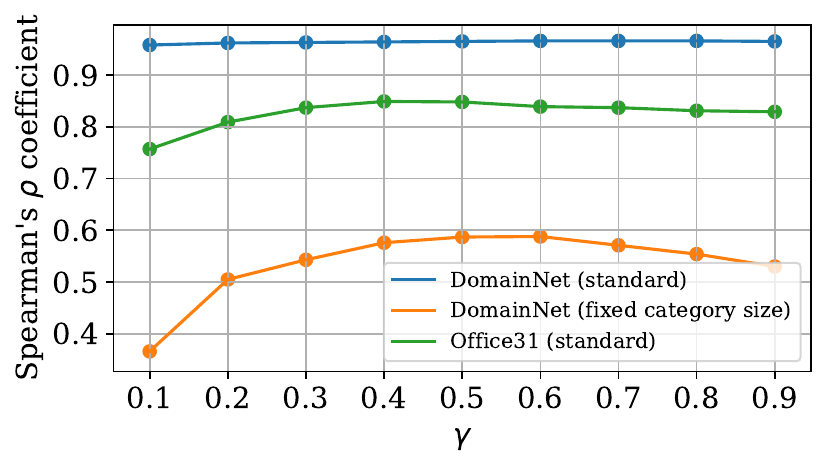}
    \vspace{-0.3cm}
    \caption{We study how the hyper parameter $\gamma$ affects the performance of JC-OTCE and find that let $\gamma = 0.5$ achieving the highest performance. }
    \label{fig:study of gamma}
\end{figure}

Here we investigate the effect of the hyperparameter $\gamma$ in JC-OTCE (see Equation (\ref{eq: cost of JC-OTCE})), which is a coefficient to balance the impacts of the sample distance and the label distance in computing the ground cost. As shown in Fig. \ref{fig:study of gamma}, the JC-OTCE metric consistently achieves the highest performance on different transfer settings when $\gamma = 0.5$.

\subsection{\revise{Application in Source Model Selection}}

\revise{One of the most straightforward applications of transferability metrics is choosing the optimal pretrained source model from a set of candidate models for a target task. In this experiment, we evaluate the effectiveness of transferability metrics in source model selection using the Visual Task Adaptation Benchmark (VTAB)~\cite{VTAB}. More specifically, the model zoo contains 15 models trained on ImageNet~\cite{russakovsky2015imagenet} by various algorithms, e.g., supervised learning (Sup-100\%) , semi-supervised learning (Semi-rotation-10\% and Semi-exemplar-10\%~\cite{zhai2019s4l}), self-supervised learning (Rotation~\cite{gidaris2018unsupervised} and Jigsaw~\cite{noroozi2016unsupervised}), generative method (Cond-biggan~\cite{brock2018large}) and VAEs~\cite{kingma2013auto}, etc. Meanwhile, VTAB provides the transfer accuracy of these models on 7 target datasets including Caltech101~\cite{Caltech101}, CIFAR-100~\cite{cifar100}, DTD~\cite{dtd}, Flowers102~\cite{flowers102}, Pets~\cite{pets}, SVHN~\cite{svhn} and Camelyon~\cite{camelyon}. In this setting, transferability metrics must identify the best source model with the highest transfer accuracy from 15 candidates. To evaluate the performance of transferability metrics in this task, we compute the Top-k (k=1,2,3) model selection accuracy, as shown in Table \ref{tab: model selection}. We observe that JC-OTCE, F-OTCE and H-score perform well in the model selection task, with JC-OTCE achieving the best accuracy. Meanwhile, in most cases, we notice that the best source model can be chosen from the predicted Top-3 highest transferable models. Note that LEEP and NCE are infeasible here since VTAB only releases the feature extractors of models.}

\begin{table}[t]
\begin{minipage}{\linewidth}



\caption{\revise{The Top-k accuracy of selecting the best source model from 15 candidate models for 7 target tasks respectively according to transferability scores.}}
\label{tab: model selection}

\centering

\begin{tabular}{l | ccc}
\toprule

Method & Top-1 & Top-2 & Top-3 \\

\midrule

JC-NCE & \textbf{3 / 7} & \textbf{5 / 7} & \textbf{6 / 7}  \\
F-OTCE & \textbf{3 / 7} & 4 / 7 & 5 / 7  \\
H-score~\cite{bao2019information} &2 / 7 & 4 / 7 &\textbf{6 / 7} \\
LogME~\cite{LogME} & 0 / 7 & 0 / 7 & 0 / 7 \\

\bottomrule

\end{tabular}

\end{minipage}

\end{table}

\subsection{\revise{Application in Multi-source Feature Fusion}}

\begin{figure}[t]
	\centering
	\includegraphics[width=0.7\linewidth]{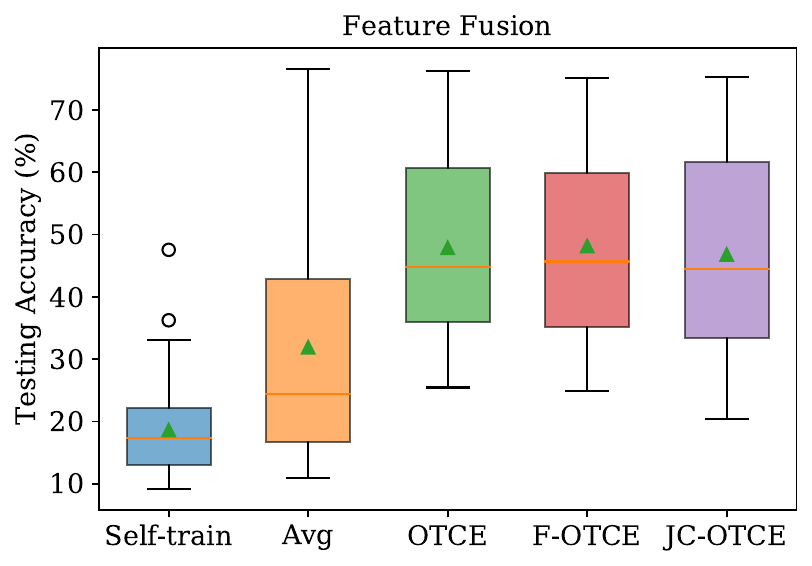}
	\caption{\revise{Testing accuracy comparisons among ``Self-train'' (directly training on target data), ``Avg'' (average fusion) and the feature fusion weighted by ``OTCE'', ``F-OTCE'', ``JC-OTCE'', respectively.}}
	\label{fig: feature fusion}
\end{figure}

\revise{When multiple source models are accessible, one can transfer them to a target task by merging their inferred features to obtain a fused feature representation~\cite{hou2017dualnet}. However, different source models may yield different transfer performances on the target task. Therefore, simple average fusion may not effectively leverage the most useful information provided by source models. As a result, we utilize transferability scores to weigh feature fusion and improve its transfer accuracy. 

We adopt the same experimental setting as proposed in \cite{OTCE}, and employ the JC-OTCE and F-OTCE scores (normalized by a softmax function) as weighting coefficients to fuse four source models trained on different domains. Specifically, we randomly sample 50 few-shot classification tasks from the \textit{Real} domain of the DomainNet dataset as target tasks. We multiply the features generated by multiple source models with the corresponding transferability-weighted coefficients and concatenate them. Then a new head classifier is trained on the fused feature representation to produce the final predictions. The results presented in Fig. \ref{fig: feature fusion} indicate that JC-OTCE and F-OTCE perform comparably to OTCE, while significantly outperforming the average fusion as expected, in the multi-source feature fusion task. 
}

\subsection{Evaluation on OTCE-based Finetune Algorithm}

One significant use of transfer learning is to address the few-shot classification problem, in which the target task has limited labeled training data, such as 1-shot or 5-shot scenarios. Earlier few-shot learning approaches~\cite{MAML, MatchingNet, ProtoNet, RelationNet} driven by meta learning only show their effectiveness in the intra-domain generalization, where the tasks used for meta-training and meta-testing are drawn from the same data distribution. Therefore, we concentrate on the more challenging few-shot classification problem across domains and tasks.

\begin{table}[t]
	\centering
\begin{minipage}{1.0\linewidth}

\centering
\setlength\tabcolsep{4pt} 

\footnotesize

\caption{Datasets used for evaluating OTCE-based finetune algorithm.  }
\label{tab: fewshot dataset}

\begin{tabular}{ccccc}
\toprule
\multirow{2} {*}{Type} & \multirow{2} {*}{Dataset} & \multirow{2} {*}{Categories} & Training & \multirow{2} {*}{Content}  \\
& & & samples & \\ 
\midrule
\multirow{2} {*}{Source} & MNIST & 10 & 60,000 & handwritten digits \\
& Caltech101 & 101& 9,146 & natural image \\
\midrule
\multirow{2} {*}{Target} & Omniglot & 1,623 & 32,460 & handwritten character \\
& MiniImageNet & 100 & 60,000 & natural image\\
\bottomrule

\end{tabular}

\end{minipage}
\end{table}

\begin{figure}[t]
\centering

\includegraphics[width=0.8\linewidth]{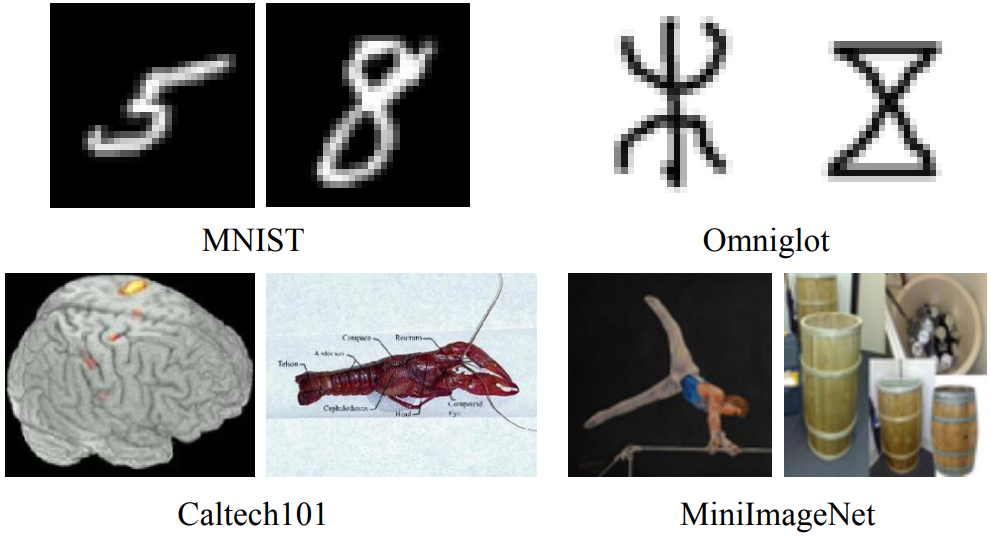}
\caption{Examples from the datasets used for model finetuning evaluations.}
\label{fig: fewshot data visualization}

\end{figure}

\begin{table}[t]
	\centering
\begin{minipage}{1.0\linewidth}

\centering
\setlength\tabcolsep{4pt} 

\footnotesize

\caption{Model architecture of the Conv4 neural network. }
\label{tab: model FewshotNet}

\begin{tabular}{cc}
\toprule
Layer name & Parameter \\
\midrule 
conv1& $3\times3$ conv, 64 filters, batch norm, ReLU, $2\times 2$ maxpooling.\\
conv2 & $3\times3$ conv, 64 filters, batch norm, ReLU, $2\times 2$ maxpooling.\\
conv3& $3\times3$ conv, 64 filters, batch norm, ReLU, $2\times 2$ maxpooling.\\
conv4& $3\times3$ conv, 64 filters, batch norm, ReLU, $2\times 2$ maxpooling.\\
\midrule
fc1 & fully connected layer, feature dim $\times$ k. \\

\bottomrule

\end{tabular}

\end{minipage}
\end{table}

\begin{table*}[t]
	\centering
\begin{minipage}{0.9\linewidth}

\centering

\footnotesize

\caption{Testing accuracy (\%) of the cross-domain cross-task few-shot classification experiments, averaged over 100 target tasks and with $95\%$ confidence intervals.}
\label{tab: otce finetune}

\begin{tabular}{clcc}
\toprule

Model & Method & MNIST $\rightarrow$ Omniglot & Caltech101 $\rightarrow$ MiniImageNet \\
\midrule
\multirow{6} {*}{Conv4} & MAML~\cite{MAML} & $88.60\pm 1.14$  & $28.23\pm 0.44 $\\
& MatchingNet~\cite{MatchingNet} &  $87.92\pm 1.10 $ & $44.75\pm 1.30 $\\
&ProtoNet~\cite{ProtoNet} & $83.11\pm 1.34 $ &  $50.40\pm 1.35 $\\
& RelationNet~\cite{RelationNet} & $69.35\pm 1.62 $ & $29.55\pm 0.61 $\\
& Vanilla finetune & $91.30\pm 0.95 $ &$49.49\pm 1.27 $ \\
& OTCE-based finetune & $\mathbf{92.32} \pm \mathbf{0.87}$ &  $\mathbf{51.36} \pm \mathbf{1.33}$ \\

\midrule
\midrule

\multirow{2} {*}{LeNet} & Vanilla finetune & $86.11\pm 1.10 $ & -\\
& OTCE-based finetune & $\mathbf{90.52} \pm \mathbf{0.94}$&- \\

\midrule
\midrule

\multirow{2} {*}{ResNet-18} & Vanilla finetune &- & $48.48\pm 1.39 $  \\
& OTCE-based finetune & -& $\mathbf{50.02} \pm \mathbf{1.34}$\\

\bottomrule

\end{tabular}

\end{minipage}
\end{table*}

\begin{table*}[t]
	\centering
\begin{minipage}{0.9\linewidth}

\centering


\caption{\revise{Classification accuracy (\%) of pretrained URL model generalizing to 100 few-shot tasks (5-way-5-shot) from each unseen domain respectively.}}
\label{tab: accuracy domain generalization}

\begin{tabular}{cccccc}
\toprule

Method & Traffic Sign & MS-COCO & MNIST & CIFAR-10 & CIFAR-100  \\

\midrule 

URL & $77.24 \pm 1.88$ & $\mathbf{72.36} \pm \mathbf{1.90}$ & $90.84 \pm 1.03$ & $67.72 \pm 1.48$ & $78.30 \pm 1.76$  \\

OTCE-based URL & $\mathbf{79.58} \pm \mathbf{1.68}$ & $71.66 \pm 1.56$ & $\mathbf{91.54} \pm \mathbf{0.96}$ & $\mathbf{68.40} \pm \mathbf{1.80}$ & $\mathbf{79.54} \pm \mathbf{1.72}$  \\

\bottomrule

\end{tabular}

\end{minipage}
\end{table*}

\textbf{Task generation}. Specifically, we generate few-shot target tasks using the character recognition dataset Omniglot~\cite{Omniglot} and the natural image classification dataset MiniImageNet~\cite{MatchingNet}, which are commonly-used benchmarks in few-shot learning. And we generate their respective source tasks using MNIST~\cite{MNIST} and Caltech101~\cite{Caltech101} datasets. The details of datasets are introduced in Table \ref{tab: fewshot dataset}, and data examples are visualized in Fig. \ref{fig: fewshot data visualization}. We randomly generate 100 few-shot image classification tasks (5-way-5-shot) from each target dataset respectively.

\textbf{Implementation details.} We train the source model (for transfer learning approaches) or apply meta-training (for meta learning approaches) on the source dataset, and then finetune the pretrained model or apply meta-test on the target task. To make a fair comparison with previous few-shot learning methods, we first evaluate performances using the widely-used Conv4~\cite{MAML, MatchingNet, ProtoNet, RelationNet} architecture, which comprises four convolutional layers and one fully connected layer, as described in Table \ref{tab: model FewshotNet}. Besides, we further examine the effectiveness of our OTCE-based finetune algorithm with different model architectures including the famous LeNet~\cite{LeNet} for character recognition (MNIST$\rightarrow$Omniglot) and the ResNet-18~\cite{he2016deep} for natural image classification (Caltech101$\rightarrow$MiniImageNet).

During the training phase of the OTCE-based finetune algorithm, we first optimize the source feature extractor over the source and target datasets for one epoch with the source batch size 256 and the target batch size 25. We use an Adam optimizer with learning rate of 0.0001. Then we initialize the target model with the optimized source weights and continue the finetuning on the target training set for 300 epochs, using the same optimizer supervised by the classification loss function. We adopt the same finetuning strategy for the vanilla finetune algorithm.

\textbf{Results.} Table \ref{tab: otce finetune} demonstrates that our proposed OTCE-based finetune method consistently improves the transfer accuracy of the vanilla finetuning under all transfer settings, with up to $4.41\%$ classification accuracy gain. On the other hand, the vanilla finetune method even outperforms existing representative few-shot learning approaches including ProtoNet~\cite{ProtoNet}, MatchingNet~\cite{MatchingNet}, MAML~\cite{MAML} and RelationNet~\cite{RelationNet} under the cross-domain setting, suggesting that current meta learning methods need further enhancements to improve their cross-domain generalization performance. \revise{Moreover, we analyze the running time of our OTCE-based finetune algorithm, and find that optimizing the F-OTCE score only accounts for 16\% of the total training time (628s) for a task pair under the Caltech101$\rightarrow$MiniImageNet setting using ResNet-18 model.}

\begin{figure}[t]
\centering
\includegraphics[width=1.0\linewidth]{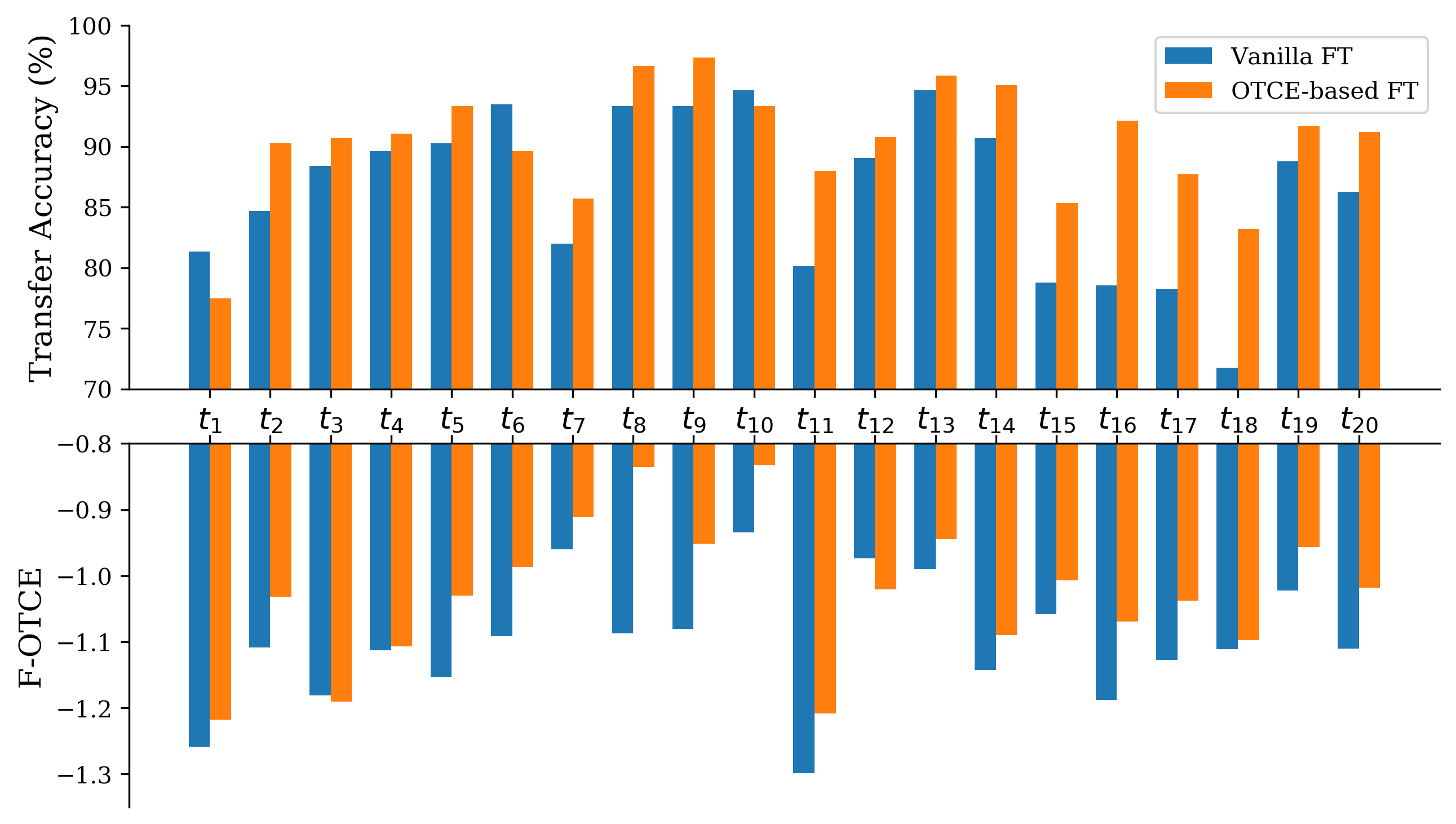}
\caption{We randomly take 20 target tasks from the \textit{MNIST to Omniglot (LeNet)} setting to visually compare the original source model (in blue) with the optimized source model (in orange) by our F-OTCE targeting to their transfer accuracy and transferability scores. It can be seen that the optimized model shows both higher transfer accuracy and higher F-OTCE score for most cases.}
\label{fig:vis trf guided TL}
\end{figure}

Fig. \ref{fig:vis trf guided TL} also visually demonstrates that our OTCE-based finetune method indeed improves both the transferability score and the transfer accuracy of the source model for most target tasks as expected. In summary, for a given target task, our OTCE-based finetune algorithm provides a straightforward yet effective approach to enhance the transferability of the source model, ultimately leading to more accurate classification results.


\subsection{\revise{Evaluation on OTCE-based URL Algorithm}}

\revise{\textbf{Dataset.} We conduct the evaluations of the OTCE-based URL algorithm on the popular Meta-Dataset~\cite{triantafillou2019meta} benchmark, which contains 8 training domains (datasets) such as ImageNet~\cite{russakovsky2015imagenet}, Omniglot~\cite{Omniglot}, Aircraft~\cite{maji2013fine}, etc, and 5 testing domains including TrafficSign~\cite{houben2013detection}, MS-COCO~\cite{lin2014microsoft}, MNIST~\cite{MNIST}, CIFAR-10~\cite{cifar100} and CIFAR-100~\cite{cifar100}. The universal model learned on training domains will be evaluated on the few-shot classification tasks randomly sampled from testing domains. 
	
\textbf{Implementation details.} We follow the default configuration of the URL algorithm, which employs ResNet-18~\cite{he2016deep} as the backbone for the universal model and domain-specific models. The universal model shares its feature extractor but not the head classifiers across domains. We optimize the F-OTCE objective using an SGD optimizer with a learning rate of 0.0001 and momentum of 0.9.

\textbf{Results.} Quantitative comparisons presented in Table \ref{tab: accuracy domain generalization} demonstrate that the generalization performances of URL incorporating with our F-OTCE metric are evidently improved, with up to 2.34\% classification accuracy gain.

}

\section{Conclusion}
\label{sec: conclusion}

Transferability estimation under the cross-domain cross-task transfer setting is a practical and challenging problem in transfer learning. Our proposed transferability-guided transfer learning framework not only provides two accurate and efficient auxiliary-free transferability metrics, F-OTCE and JC-OTCE, without the need of retraining the source model, but also offers an useful objective function (F-OTCE) to enhance the generalizability of the source model, ultimately leading to higher transfer accuracy in downstream model finetuning and domain generalization tasks. F-OTCE is computed as the negative conditional entropy between the source and target labels when given the optimal coupling produced by Optimal Transport. The conditional entropy measures the predicted label uncertainty of the target task under the given pretrained source model and source data, which is negatively correlated with the ground-truth transfer accuracy such that it can serve as an accurate indicator of transferability. Furthermore, JC-OTCE includes the additional label distance in building more accurate data correspondences, which trade-offs a minor efficiency drop for more accurate transferability estimation under diverse transfer configurations.

Our proposed F-OTCE and JC-OTCE metrics drastically reduce the computation time of the auxiliary-based OTCE from 43 minutes to 9.32s and 10.78s respectively, while consistently showing higher accuracy in predicting ground-truth transfer performance than state-of-the-art auxiliary-free metrics, achieving average correlation gains of $21.1\%$ and $25.8\%$ respectively. In particular, JC-OTCE performs comparably to the original OTCE in transferability accuracy, with greater flexibility and efficiency. Additionally, our transferability-guided transfer learning algorithms improves the transfer accuracy of the source model with up to $4.41\%$ and $2.34\%$ gains in few-shot classification tasks. We believe that our OTCE framework can inspire various downstream tasks in transfer learning, multi-task learning and other related applications.

\section{Acknowledgement}
\label{sec: acknowledgements}
This work is supported in part by the Natural Science Foundation of China (Grant 62371270) , Tsinghua SIGS Scientific Research Start-up Fund (Grant QD2021012C) and  Shenzhen Key Laboratory of Ubiquitous Data Enabling (No.ZDSYS20220527171406015).

\bibliographystyle{IEEEtran}
\bibliography{IEEEabrv,reference}

\vfill

\end{document}